\RequirePackage{fix-cm}
\documentclass[twocolumn]{svjour3}          
\smartqed  
\usepackage{graphicx}
\usepackage{soul}
\usepackage{cite}
\usepackage{amsmath,amssymb,amsfonts}
\usepackage{algorithmic}
\usepackage{graphicx}
\usepackage{textcomp}
\usepackage{comment}
\usepackage{soul}
\usepackage{color,xcolor}

\usepackage{caption}
\usepackage{subfigure}
\begin{document}

\title{ACFNet: Adaptively-Cooperative Fusion Network for RGB-D Salient Object Detection
}


\author{Jinchao Zhu$^{1, 2}$ \and Xiaoyu Zhang*$^{1, 2}$ \and Xian Fang$^{3}$  \\
\and Feng Dong$^{1}$ \and Yan Sheng$^{1}$ \and Siyu Yan$^{1}$ \and Xianbang Meng$^{1}$
}


\institute{1 College of Artificial Intelligence, Nankai University, Tianjin, China. \\
2 Tianjin Key Laboratory of Intelligent Robotics, Nankai University, Tianjin, China.\\
3 College of Computer Science, Nankai University, Tianjin, China.
}

\date{Received: date / Accepted: date}

\maketitle

\begin{abstract}
The reasonable employment of RGB and de-pth data show great significance in promoting the development of computer vision tasks and robot-environment interaction.
However, there are different advantages and disadvantages in the early and late fusion of the two types of data.
Besides, due to the diversity of object information, using a single type of data in a specific scenario tends to result in semantic misleading.
Based on the above considerations, we propose an adaptively-cooperative fusion network (ACFNet) with ResinRes structure for salient object detection.
This structure is designed to flexibly utilize the advantages of feature fusion in early and late stages.
Secondly, an adaptively-cooperative semantic guidance (ACG) scheme is designed to suppress inaccurate features in the guidance phase.
Further, we proposed a type-based attention module (TAM) to optimize the network and enhance the multi-scale perception of different objects.
For different objects, the features generated by different types of convolution are enhanced or suppressed by the gated mechanism for segmentation optimization.
ACG and TAM optimize the transfer of feature streams according to their data attributes and convolution attributes, respectively.
Sufficient experiments conducted on RGB-D SOD datasets illustrate that the proposed network performs favorably against 18 state-of-the-art algorithms.
\end{abstract}


\section{Introduction}
For the interaction between robots and the environment, only using RGB image information is easy to form misjudgment, so depth information is an important tool for robots.
The maximum use of depth images will greatly improve the salient object detection (SOD) results and promote the development of human-computer interaction.
With the progress of sensor accuracy and the maturity of industrial technology, some off-shelf RGB-D sensors have been widely adopted by robot systems.
This trend will be more pronounced as the cost of depth sensors decreases.
Compared with RGB image information, depth information is more robust in some specific scenes.
Therefore, using depth information and RGB data at the same time is of great significance to enhance the robot's perception of the environment.

Salient object detection simulates human visual attention to find salient object areas in the field of vision.
It has been widely used in computer vision tasks, such as pedestrian Re-ID~\cite{2013Unsupervised}, foreground analysis~\cite{2018-IJCAI-Em}\cite{2017-ICCV-Sm}, semantic segmentation~\cite{Wei2017STC} and so on.
The early SOD methods are mainly based on low-level features and hand-crafted features~\cite{2009-CVPR-Fm}\cite{2015-PAMI-Global}\cite{2013--CVPR-Submodular}, but the design of these features requires strong prior knowledge and the generalization ability of the model is weak.
With the development of deep learning, deep features with strong semantic information are used for multi-level feature fusion, which greatly improves model performance.

\begin{figure}[t]
\centering
\includegraphics[width=1\columnwidth]{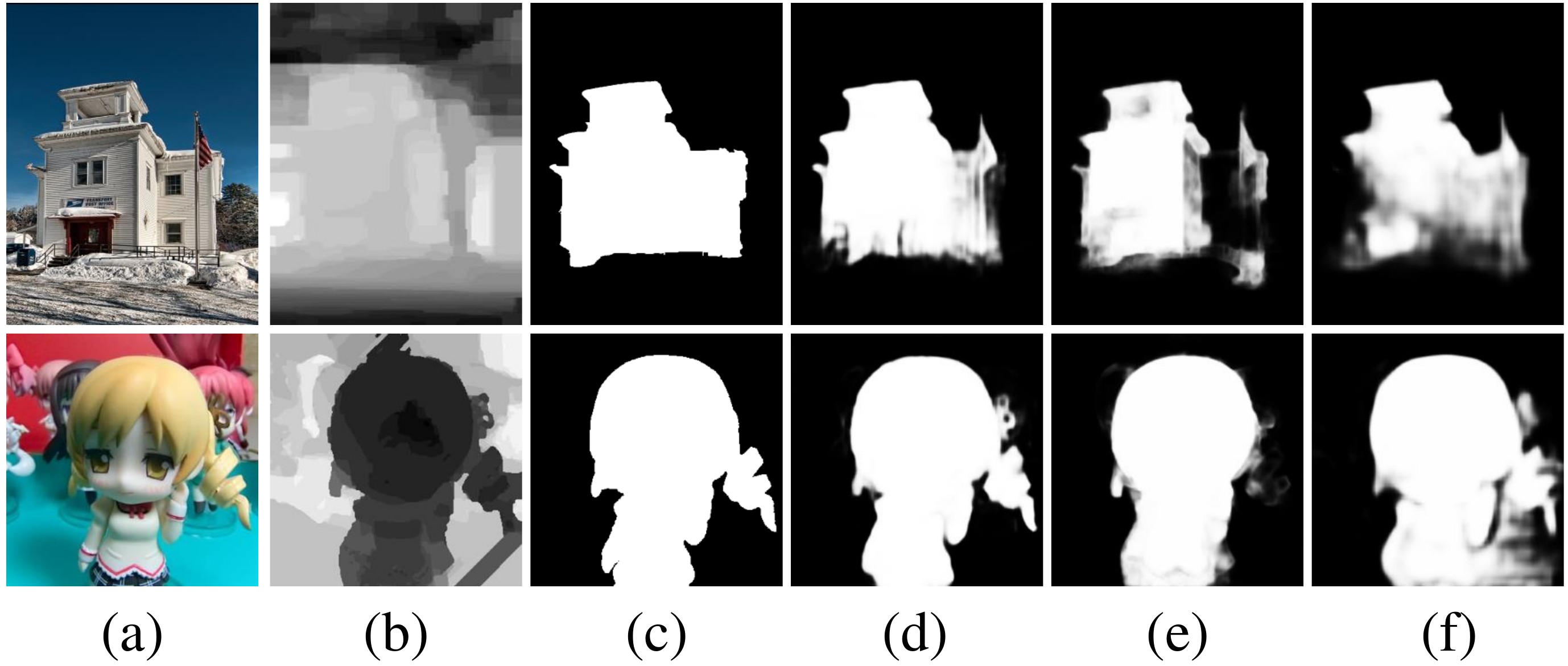} 
\caption{Visual comparison of our method and other
approaches. (a) RGB (b) Depth (c) GT (d) Ours (e) S2MA~\cite{2020-CVPR-S2MA} (f) D3Net~\cite{2020-TNNLS-D3Net}.
\textbf{Given depth images like (b), how to balance the influence of depth data and RGB data to accurately detect salient objects?}
The depth information of the first row is easy to lead to misjudgment, so that the right half of the house (e) is lost .
The depth data in the 2nd row is helpful to accurately identify the right half of the doll, which is not fully utilized by (e) and (f).
}
\label{F1}
\end{figure}

RGB images have rich color data which includes the appearance information and texture details of the objects.
But sometimes the color information can be deceptive and lead to the misjudgment of the object's environment.
Depth images have robust structural information in certain situations.
However, depth images are not sensitive to the internal details on the plane, as shown in Fig.\ref{F1}.
Therefore, excellent algorithms~\cite{2019-TIP-TANet}\cite{2019-CVPR-CPFP}\cite{2019-ACCESS-AFNet} based on the two kinds of data are constantly proposed.  

In order to better introduce our algorithm, according to the fusion structure, we follow~\cite{2020-RGBD-survey}\cite{2017-IROS-M3Net} to roughly divide the existing network into three main structures:
input fusion~\cite{M3Net-22}\cite{M3Net-24}\cite{2020-ECCV-DANet},
early fusion\cite{2017-IROS-M3Net},
late fusion\cite{M3Net-7}\cite{M3Net-8}\cite{M3Net-10}\cite{2020-CVPR-S2MA}\cite{2021-TMM-cmSalGAN},
and multi-scale fusion~\cite{2020-TIP-DPANet}~\cite{2021-CVPR-DCF} (as shown in Fig.\ref{early} (a) (b) (c) (d)).
Most papers pay more attention to the encoder stage for the classification of fusion methods.
According to~\cite{2020-RGBD-survey}, we further divide the late fusion into later fusion~\cite{2020-CVPR-S2MA} and late result fusion~\cite{2021-TMM-cmSalGAN} as shown in Fig.\ref{early} (c).


In general, the early fusion structure can comprehensively utilize the rich details of the two kinds of low-level features, but the noise of low-level features, the unhelpful data in special cases (especially the depth images), and the lack of semantic information may mislead deeper features.
The late fusion structure can better judge the validity of the two types of input at the semantic level, but due to the different environments of salient objects, sometimes the advantages of comprehensive inferring based on lower-level features are weakened.
Therefore, $\textit{how}$ $\textit{to}$ $\textit{dynamically}$ $\textit{balance}$ $\textit{the}$ $\textit{advantages}$ $\textit{of}$ $\textit{early}$ $\textit{fusion}$ $\textit{and}$ $\textit{late}$ $\textit{fusion}$ is the key issue of our research.
In addition, the valueless depth images (depth data in the 9th row of Fig.~\ref{compare}) or the undecidable nature of some RGB images (RGB data in the 2nd row of Fig.~\ref{compare}) can also have a serious negative impact on the results.  
\cite{2020-TNNLS-D3Net} proposed a depth depurator unit (DDU) to determine whether to use RGB-D generated results or RGB only generated results (Fig.\ref{early} (a)).
But it directly uses a whole U-shaped network (the 3rd network in Fig.\ref{early} (a)) to evaluate the reliability of depth information.
We think the features in the third network are wasted.
Because the depth image sometimes has local validity, as shown in Fig.\ref{pipline} butterfly (butterfly's right contour information) and Fig.\ref{compare} (local area on the left side of lighter).
$\textit{How}$ $\textit{to}$ $\textit{fully}$ $\textit{exploit}$ $\textit{the}$ $\textit{depth}$ $\textit{information}$ $\textit{and}$ $\textit{suppress}$ $\textit{invalidity}$ $\textit{features}$ is another key issue of our research.

In this paper, we design an adaptively-cooperative fusion network (ACFNet) based on RGB information semantic guidance and depth information semantic guidance.
Firstly, we design a network structure of three encoders and three decoders, which has the advantages of early fusion and late fusion.  
We not only send the stage features (R2',R3',R4',D2',D3',D4') of Decoder1-R and Decoder1-D in Fig.\ref{pipline} to Decoder2-F but also use the final outputs of their respective decoders as inputs for strong semantic guidance.
Because the final outputs of the first two ResNets (FPN outputs of Encoder1-R and Encoder1-D) are the inputs of the 3rd ResNet and the combination of the FPN outputs and the stage outputs form the residual structure at the macro level, this structure is named ResinRes.
Secondly, in order to avoid the inaccurate guidance of strongly semantic information of RGB features or depth features, a gate unit is designed to evaluate the semantic credibility to suppress the bias guidance.
Finally, a type-based attention module is adopted to further enhance the final output of each FPN.
Our contributions are summarized as follows:
\begin{itemize}
\item[(1)] We propose a ResinRes network structure, which flexibly utilizes the advantages of early and late fusion.
A adaptively-cooperative semantic guidance (ACG) scheme is designed to facilitate the adaptive cooperation of different semantic features.
	
\item[(2)] An type-based attention module (TAM) is proposed to optimize features and enhance the perception of different scale objects. ACG and TAM optimize the transfer of feature streams according to the data attribute and convolution attribute of features respectively.
	
\item[(3)] Sufficient experiments conducted on 8 RGB-D SOD datasets illustrate that our model performs favorably against 16 state-of-the-art algorithms.
To prove the universality of the proposed model, we also verified the model effect on an RGB-NIR dataset for the multi-spectral SOD task.
The model performs favorably against 6 state-of-the-art algorithms for RGB-NIR SOD.
\end{itemize}

\section{RELATED WORK}
\subsection{Salient Object Detection with Multiple Cues}
Early salient object detection (SOD) is mainly based on hand-crafted features and low-level features, such as color contrast~\cite{2009-CVPR-Fm}, center prior~\cite{2013--CVPR-Submodular}, etc.
\begin{figure*}[t]
\centering
\includegraphics[width=2\columnwidth]{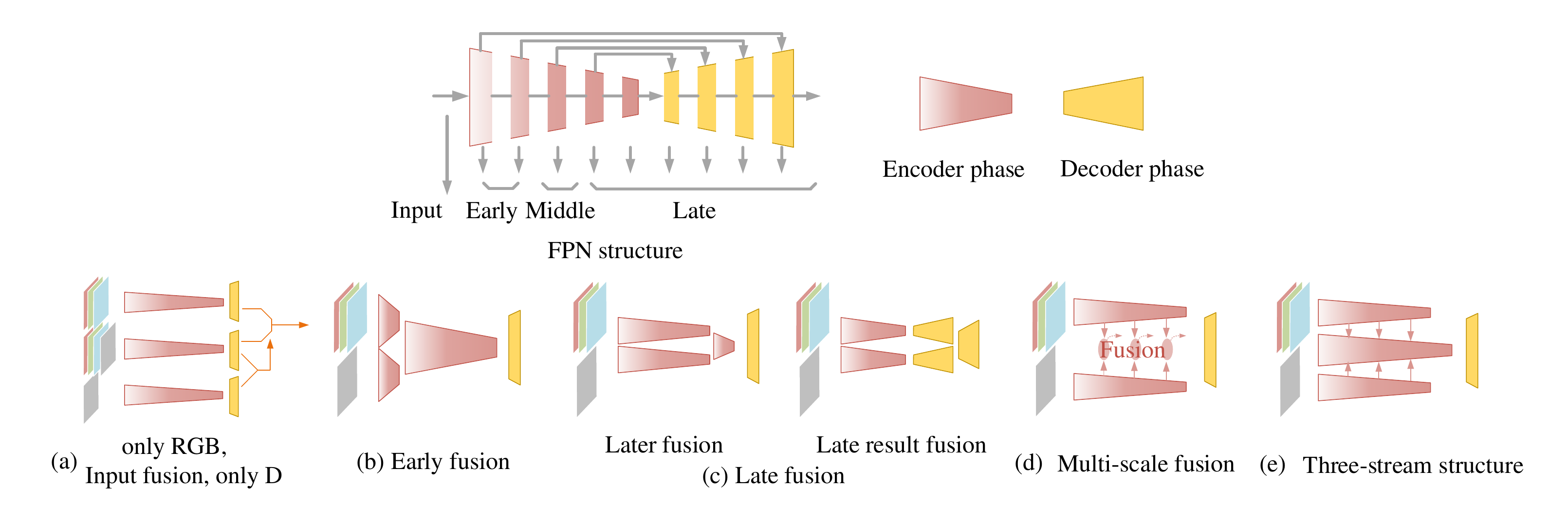} 
\caption{Different network architecture design.
We define the output features of each stage in the FPN structure.
On the left side of each picture, red, green, and blue squares represent RGB images and gray squares represent depth maps.
The red areas in the middle represent the encoders.
The yellow parts on the right represent the decoder and output.
The orange arrow on the right side of figure (a)~\cite{2020-TNNLS-D3Net} indicates whether to use the 1st or 2nd network of figure (a) according to the output similarity of the 2nd and 3rd network.
The 3 networks in figure (a) represent different inputs: RGB, RGB-D (input fusion), and D.
(b), (c) and (d) represent early fusion, late fusion, and multi-scale fusion respectively. (e) is a special case of (d).
}
\label{early}
\end{figure*}
With the development of deep learning, convolutional neural networks using multi-scale features have become the dominant method~\cite{2020-AAAI-F3Net}\cite{2020-AAAI-GCPANet}\cite{2019-CVPR-PAGE}.
The U-shaped network represented by FPN~\cite{2027-CVPR-FPNdetection} and UNet~\cite{2015-ICM-Unet} has become the current mainstream structure.
In order to further obtain a more accurate pixel-level prediction of salient objects, scholars use a variety of cues to train better models.
These cues include edge cue~\cite{2019-ICCV-EGNet}\cite{2020-APID-Bay}, eye fixation cue\cite{2019-PAMI-ASNet}, central region cue~\cite{2020-CVPR-LDF}, multi-spectral cue~\cite{2020-TMM-RGBTSOD}, and depth information cue~\cite{2019-ACCESS-AFNet}.
\cite{2019-ICCV-EGNet} used edge guidance to improve algorithm performance.
\cite{2019-ICCV-SCRN} constructed a complementary promotion model, in which edge detection and salient object detection complement each other.
Both \cite{2019-CVPR-MLMSNet} and \cite{2019-CVPR-PoolNet} used standard edge datasets and saliency datasets for joint training.
\cite{2019-PAMI-ASNet} used eye fixation labels to infer salient objects and the advantage of eye fixation cue was fully exploited by LSTM (Long Short-Term Memory).
\cite{2020-CVPR-LDF} designed labels that focus on body information and detail information respectively to identify the regions of salient objects from different perspectives.

With the development of sensors collecting depth information, depth data has become an important cue to improve the saliency detection effect.
A large number of outstanding algorithms~\cite{2019-TIP-TANet}\cite{2019-CVPR-CPFP}\cite{2018-CVPR-PCANet}\cite{2019-ACCESS-AFNet} for salient object detection based on RGB-D images have been proposed. 
As a representative method of input fusion, \cite{2020-ECCV-DANet} uses a depth-enhanced dual attention mechanism to strengthen the mask-guided attention.
\cite{2017-IROS-M3Net} designed a multi-path multi-modal fusion method to combine the advantages of early and late fusion.
\cite{2019-TIP-TANet} proposed a three-stream structure (Fig.\ref{early} (d)), in which the features of RGB branch and depth branch were fused to the middle branch in each stage.
Besides, \cite{2020-ECCV-BBSNet} sends the multi-level enhanced depth information to the RGB branch level by level.
They are two special cases of multi-scale fusion.
As a representative method of late result fusion, \cite{2021-TMM-cmSalGAN} interactively mixes the two modal features in the decoder output stage by generative adversarial representation learning.

%


\begin{figure*}[htb]
	\centering
	\includegraphics[width=2\columnwidth]{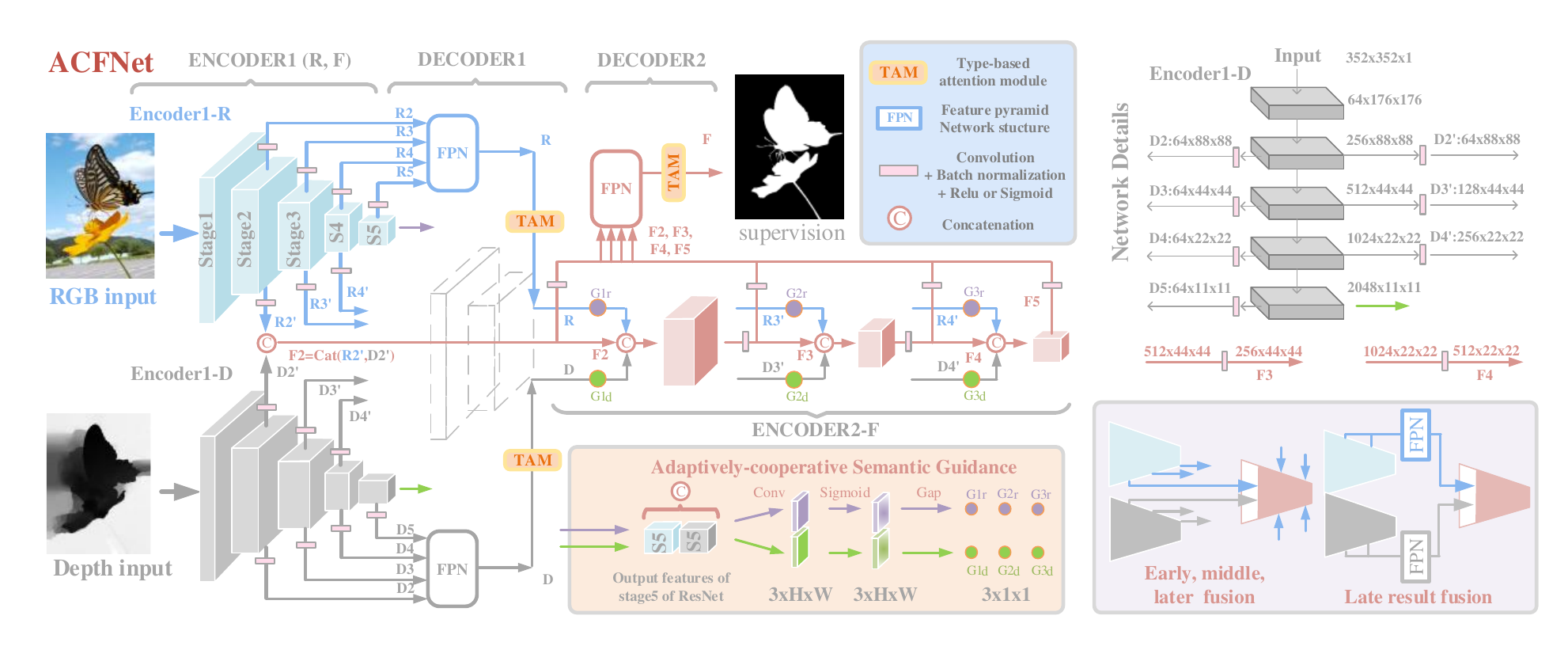} 
	\caption{Adaptively-cooperative fusion Network.
		Encoder1 consists of two encoders, which are RGB image encoder (Encoder-R) and depth image encoder (Encoder-D).
		Decoder1 is composed of two feature pyramid networks (FPN), which integrate two kinds of multi-level features respectively.
		In the 2nd phase, the Encoder2-F takes the outputs of decoder 1 and encoder 1 as the inputs.
		The gate unit of ACG uses the outputs of the 5th stage of Encoder-R and Encoder-D to generate weights, which are used to adaptively regulate Encoder2-F.
		Decoder2 is also a feature pyramid network.
		After the output of each decoder, there is a type-based attention module (TAM) for feature enhancement.
		All encoders are ResNet-50.
		Stage 1,2 of Encoder2-F are removed.
		The top right corner shows the details of the network.
		In the lower right corner, we analyze the various fusion structures of ACFNet.
	}
	\label{pipline}
\end{figure*}

\subsection{Attention Mechanisms and Gated Mechanisms}

Attention mechanism and gated mechanism play an important role in the optimization of convolutional neural networks.
\cite{2018-CVPR-SENet} greatly improved the quality of feature representation by establishing the interdependence between convolution feature channels.
After channel attention~\cite{2018-CVPR-SENet} is proposed, \cite{2018-ECCV-CBAM} designed spatial attention to further strengthen features.
These ideas have been widely adopted in the field of SOD.
\cite{2019-TIP-TANet} proposed Att-CMCL block to reduce cross-modal cross-level fusion uncertainty and promoted the fusion effect by attention mechanism.
The DMSW module of \cite{2019-ICCV-DMRA} adopted dilated convolution to increase the size of the receptive field and used the attention mechanism to enhance the information of the useful channels. 

Besides, the advantages of the gated mechanism in \cite{2020-ECCV-GateNet}\cite{2018-CVPR-BMPM} were fully demonstrated and the network was balanced and optimized.
\cite{2020-TNNLS-D3Net} used a unique gate unit (DDU), and MAE (mean absolute error) metric was adopted to compare the results of RGBD network and depth network to determine whether the outputs of RGB network or RGBD network are adopted (Fig.\ref{early} (a)).
Compared with \cite{2020-TNNLS-D3Net}, our gate unit (ACG) works inside the network, and the gate adjustment range is 0 to 1 instead of 0 or 1.
We make full use of the depth information rather than directly abandon the results of the depth network.
Our gate unit is unsupervised and its parameters are formed spontaneously with the training process.
The gate unit is simple and easy to implement and does not need to design complex supervision labels.
\cite{2020-arxiv-detectRS} proposed a switchable atrous convolution (SAC) to detect objects of different scales by dynamically switching convolutions of multiple dilation ratios, in which the gated mechanism plays an important role.
Inspired by this idea, we design a type-based attention module to analyze and regulate the contribution of features from various types of convolution.

\section{PROPOSED METHOD}
In this section, We first introduce the design motivation of the ResinRes network structure.
Then we explore the implementation of adaptively-cooperative semantic guidance (ACG) scheme.
Finally, we discuss the details of the type-based attention module (TAM).
\subsection{ResinRes Structure}
In the previous work, input fusion, early fusion, late fusion, and other fusion schemes have been explored.
Early fusion is helpful to promote comprehensive inferring of low-level features.
However, late fusion comprehensively judges the credibility of two types of data at the semantic level,
which helps to prevent the local noise of individual low-level features from being amplified at the semantic level.
Therefore, we design a ResinRes structure to be compatible with the advantages of the two designs.
Besides, it is worth noting that we use weighted strong semantic information (R, D) for low-level feature (R2', D2') semantic guidance, and adopt gate units to balance the influence of the two guidance.
Most of the previous work design a complex decoder to fuse depth and RGB features.
However, we adopt the 3rd backbone (ResNet50) to re-encode multimodal features and achieve better feature fusion.

Specifically, the ResinRes structure can be divided into two phases.
The first phase consists of two encoders (Fig.\ref{pipline} Encoder1-R, Encoder1-D) and two decoders (feature pyramid networks).
The second phase consists of an encoder (Fig.\ref{pipline} Econder2-F) and a decoder (FPN).
Encoder1-R and Encoder1-D only encode and decode their input data (single mode) to get output R and D.
After feature integration of FPN, the output features (R, D) with rich semantic information are sent to Encoder2-F as strong semantic guidance information streams of RGB data and depth data respectively.
The late result fusion diagram in the lower right corner of Fig.\ref{pipline} shows the formation process of these two information streams.

The side outputs $R2'$ and $D2'$ between Encoder1-D and Encoder1-R in Fig.\ref{pipline} are fed to phase 2 as early fusion features.
Encoder2-F makes comprehensive inferring based on these mixed low-level features.
The side outputs $R3'$, $R4'$, $D3'$, $D4'$ are fused into the input of each stage of Encoder2-F.
We call them middle fusion ($R3'$, $D3'$) and later fusion ($R4'$, $D4'$). 
The lower right corner of Fig.\ref{pipline} shows the structure of early, middle, and later fusion.
It is worth noting that the low-level features $R2'$, $D2'$, and the final outputs $R$ and $D$ of ResNet-50s form a macro residual structure.
Besides, they are sent together into another ResNet-50 encoder, so we call this structure ResinRes.
Because of the structural symmetry of ACFNet, we only show the details of Encoder1-D on the right of Fig.\ref{pipline}.
In Encoder2-F, we remove stage 1 and stage 2 (the dotted line on the left side of Encoder2-F) of ResNet-50 to simplify the network.

\subsection{Adaptively-Cooperative Semantic Guidance}
The ResinRes structure synthesizes multiple fusion structures. Therefore, it is also a special case of multi-scale fusion.
In this section, we further design an adaptively cooperative semantic guidance (ACG) scheme to dynamically balance their advantages (early fusion (EF), middle fusion (MF), and late fusion (LF)),  fully exploit the depth information, and suppress invalidity features.
It is worth noting that we use the unsupervised gate unit to assist multi-modal feature fusion.
Our gate unit is simple and easy to implement, and the parameter learning of the ACG module is formed spontaneously with the training process.
Compared with~\cite{2020-TNNLS-D3Net}, our ACG module suppresses and balances features within the network, rather than selecting final output results.

Specifically, $R$ and $D$ are features with rich semantic information generated by a single type of data.
They are adopted to guide the low-level feature $R2'$, $D2'$ in phase 2.
Similar to $R$ and $D$ (semantic guidance of different modalities), $R3'$, $R4'$, $D3'$, $D4'$, the side outputs of ResNet-50, are used for encoding materials in later stages.
RGB features ($R3'$, $R4'$), depth features ($D3'$, $D4'$) and hybrid features ($F3$, $F4$) cooperate adaptively under the regulation of the ACG.

Considering the local inefficiency of depth data (the left side of the butterfly in Fig.\ref{pipline}) and the local ambiguity of RGB data (the 1st row in Fig.\ref{compare}) in special cases, we propose the ACG module to evaluate the credibility of single modal semantic information.
ACG synthetically compares the deepest features of Encoder1-R and Encoder1-D to allow the network to learn the weight of each guidance position.
This process is similar to the classification prediction of input data in a classification network.
The difference is that our prediction result is not the probability of a certain category, but the credibility of depth features and RGB features.

The specific process is as follows:
\begin{small}
	\begin{align}\label{E-G}
		G1_{r}, G2_{r}, G3_{r} = G_{r} = Gap(S(C^{3}(Cat(S5_{r}, S5_{d}))))\\
		G1_{d}, G2_{d}, G3_{d} = G_{d} = Gap(S(C^{3}(Cat(S5_{r}, S5_{d}))))
	\end{align}
\end{small}  
where $S5_{r}, S5_{d}$ are the outputs of stage 5 of Encoder1-R and Encoder1-D, respectively.
$G_{r}$ and $G_{d}$ correspond to the purple and green dots in Fig.\ref{pipline} ACG.
$Cat(\cdot)$ stands for concatenation.
$C^{3}(\cdot)$ represents convolution and batch normalization operations and it transforms the input feature into 3 channels.
$S(\cdot)$ is sigmoid.
$Gap$ is global average pooling.
Fig.\ref{pipline} shows the change process of feature shape (4096$\times$H$\times$W$\rightarrow$3$\times$H$\times$W$\rightarrow$3$\times$1$\times$1).

The input of each stage of Encoder2-F can be expressed as:
\begin{small}
	\begin{align}\label{E-ACG}
		I1 &= Cat(F2*1, R*G1_{r}, D*G1_{d})\\
		I2 &= Cat(F3*1, R3'*G2_{r}, D3'*G2_{d})\\
		I3 &= Cat(F4*1, R4'*G3_{r}, D4'*G3_{d})
	\end{align}
\end{small}
When $G1_{r}=G2_{r}=G3_{r}=G1_{d}=G2_{d}=G3_{d}$ = 0, the network can be considered as early fusion.
When $G1_{r}=G1_{d} = 1, G2_{r}=G2_{d}=G3_{r}=G3_{d} = 0$, the late result fusion structure plays a leading role.
Similarly, we can deduce the situation when middle fusion or later fusion is the dominant structure.
The above are extreme cases.
We introduce them in detail in the supplementary materials.

\subsection{Type-Based Attention Module}
To further improve the network performance, we use type-based attention modules (TAM) to enhance the output of each decoder (FPN).
The widely used feature pyramid network (FPN) architecture is used to integrate the multiple output features of the encoder.
Taking the 2nd phase decoder as an example, the specific process is as follows:
\begin{small}
	\begin{align}\label{E-FPN}
		&F4_{fpn} = C_{3\times3}(F4 + U(F5))\\
		&F3_{fpn} = C_{3\times3}(F3 + U(F4_{fpn}))\\
		&F = C_{3\times3}(F2 + U(F3_{fpn}))
	\end{align}
\end{small}
$C_{3\times3}(\cdot)$ denotes convolution, batch normalization and ReLU operations.
$U(\cdot)$ means upsampling.

\begin{figure}[t]
	\centering
	\includegraphics[width=1\columnwidth]{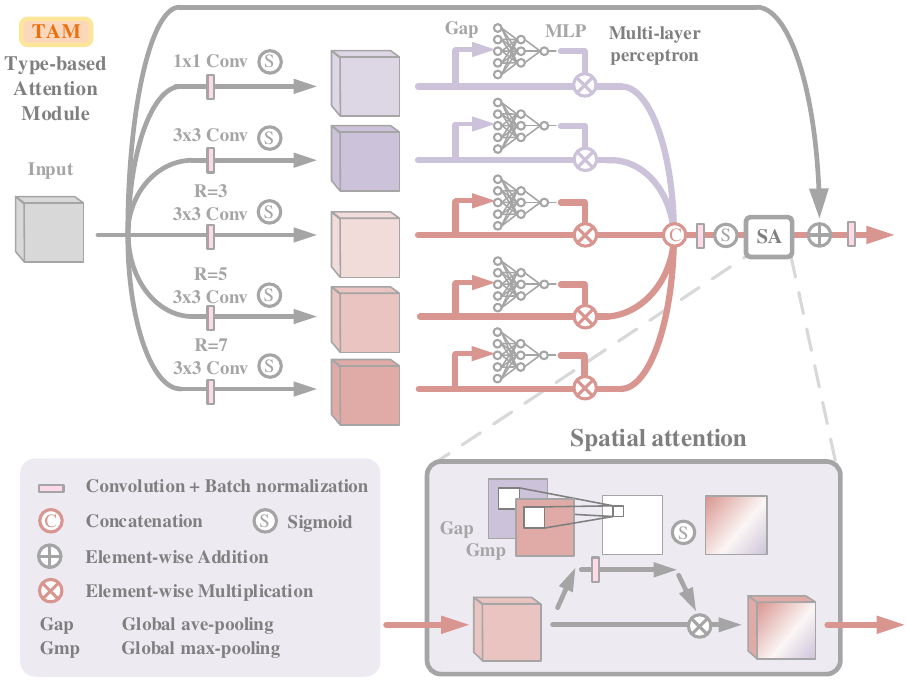} 
	\caption{Type-based attention module.
	}
	\label{TAM}
\end{figure}

The dilated convolution has a larger receptive field than normal convolution,
and the mixture of convolutions with various dilation rates~\cite{2018-TPAMI-ASPP}\cite{2018-ECCV-RFB} is beneficial to the perception of salient objects of different scales~\cite{2020-arxiv-detectRS}.
Inspired by the SAC module of~\cite{2020-arxiv-detectRS}, we design a type-based attention module (TAM) to maximize the ability of different types of convolution in different scenarios, as shown in Fig.\ref{TAM}.
We design a multi-layer perceptron for each type of convolution to evaluate the contribution of each type of feature.
The traditional channel attention module~\cite{2018-CVPR-SENet} recalibrates each channel of the features, while TAM recalibrates features uniformly according to the type of feature.
The specific process can be expressed by the following equations:
\begin{small}
	\begin{align}\label{E-TAM}
		&F_{1\times1}, F_{3\times3} = C_{1\times1}(I), C_{3\times3}(I)\\
		&F_{3\times3-3}, F_{3\times3-5}, F_{3\times3-7} = C_{3\times3-3}(I), C_{3\times3-5}(I), C_{3\times3-7}(I)\\
		&G = MLP(Flatten(Gap(F)))\\
		&F_{TA} = C^{64}_{3\times3}(Cat(F_{1\times1}*G1,F_{3\times3}*G2,\\
		&F_{3\times3-3}*G3,F_{3\times3-5}*G4,F_{3\times3-7}*G5))
	\end{align}
\end{small}
Where $I$ is the input feature.
$C_{3\times3-3}(\cdot)$ denotes convolution, batch normalization and sigmoid operations.
${3\times3-3}$ represents $3\times$3 convolution and dilation ratio is 3.
$F_{3\times3-3}$ represents the feature obtained by $C_{3\times3-3}$.
Gap is the global average pooling in the spatial dimension.
The shape of Gap(F) is $64\times1\times1$.
$Flatten(\cdot)$ is flatten operation.
$MLP$ is a multi-layer perceptron composed of fully-connected layers.
The numbers of neurons in three layers (MLP) are 64-4-1.
To save space, we use Eq.11 to represent the calculation process of G1, G2, G3, G4, G5.
The superscript of $C^{64}_{3\times3}$ means that the input feature is transformed into 64 channels by a convolution operation.

After reallocating weights to the features obtained by different types of convolution, we use the spatial attention module~\cite{2018-ECCV-CBAM} to enhance the expression ability of features at the spatial dimension. The process is as follows:
\begin{small}
	\begin{align}\label{E-SA}
		&F_{SA} = F_{TA}\times C^{1}_{3\times3}(Cat(Gap_{c}(F_{TA}), Gmp_{c}(F_{TA}))) \\
		&output = C_{3\times3}(I + F_{SA})
	\end{align}
\end{small}
It is worth noting that $Gap_{c}$ and $Gmp_{c}$ are global ave-pooling and global max-pooling on the channel axis, respectively.

\section{SUPERVISION}
Binary cross entropy (BCE) loss and a IoU loss are adopted to supervise the outputs.
\begin{equation}\label{E-loss}
	L = L_{bce} + L_{iou}
\end{equation}
Binary cross entropy loss is widely adopted in segmentation algorithms.
BCE loss is defined as:
\begin{small}  
	\begin{equation}\label{E-BCE}
		L_{bce} = - \sum_{(x,y)}[G(x,y)log(P(x,y))+(1-G(x,y))log(1-P(x,y))]
	\end{equation}
\end{small}
where $P(x,y) \in [0,1]$ is the prediction value of saliency map at pixel $(x,y)$.
$G(x,y) \in [0,1]$ is the ground truth (GT) label of the point $(x,y)$.

IoU loss evaluates the similarity between the prediction result and GT from a holistic perspective rather than from a single pixel.
\begin{small}
	\begin{equation}\label{E-IOU}
		L_{iou}=1 - \frac{\sum_{(x,y)}[G(x,y)*P(x,y)]}  {\sum_{(x,y)}[P(x,y)+G(x,y)-G(x,y)*P(x,y)] }\\
	\end{equation}
\end{small}

\section{EXPERIMENTS}
\subsection{Datasets and Evalution Merics}
We verify the effectiveness of our proposed algorithm on 8 datasets.
\textbf{NJUD}~\cite{2015-SPIC-NJUD} contains 1,985 RGB images with corresponding label and depth images, which are from Internet and stereo camera. 
\textbf{NLPR}~\cite{2014-LNCS-NLPR} (RGBD1000) contains 1,000 RGBD images obtained by Kinect. Some images have multiple salient objects. 
\textbf{RGBD135}~\cite{2014-ICIMCS-RGBD135} contains 135 RGBD images captured by Kinect and the dataset is also named DES. 
\textbf{SIP}~\cite{2020-TNNLS-SINet} consists of 1,000 persion RGBD images captured by Huawei Meta10. 
\textbf{SSD}~\cite{2017-ICCVW-SSD} consists of 80 images picked up from stereo movies.  
\textbf{STEREO}~\cite{2012-CVPR-STEREO} (another name is SSB) contains 1,000 stereoscopic images from Internet. 
\textbf{DUTRGBD}~\cite{2019-ICCV-DUTRGBD} contains 1,200 RGBD images with indoor and outdoor scenes. It is split into two parts for training (800) and testing (400).  
\textbf{LFSD}~\cite{2017-TPAMI-LFSD}  contains 100 RGBD images captured by light field camera.  

We follow~\cite{2019-ICCV-DMRA} to design the experimental scheme. 
For DUTRGBD, we use 800 pairs of pictures for training and 400 pairs for testing.
For the other 7 datasets, we follow the dataset partition of \cite{2018-CVPR-PCANet}\cite{2018-PR-MMCI}\cite{2017-IROS-M3Net}. 
1,485 pairs of pictures sampled from NJUD and 700 pairs of pictures sampled from NLPR are used for training dataset and the remaining images are used as testing datasets.

We used six widely used metrics to evaluate the performance of our model and the state-of-the-art methods. 
Mean absolute error~\cite{2012-CVPR-MAE} (\textbf{MAE}) are adopted to estimate the pixel-level approximation degree between GT and the prediction.
\textbf{S-measure}~\cite{2017-ICCV-Sm} ($\textbf{S}_{m}$) estimates the structure similarities by equation: $S_{m} = \alpha\cdot S_{o} + (1-\alpha)\cdot S_{r}$, $\alpha=0.5$.  
\textbf{E-measure}~\cite{2018-IJCAI-Em} ($\textbf{E}_{m}$) are also widely used in previous works.  
\textbf{PR curve} compares the prediction results and GT to calculate the precision ($TP/(TP+FP)$) and recall ($TP/(TP+FN)$).
\textbf{F-measure} ($\textbf{F}_{\beta}$)~\cite{2009-CVPR-Fm} use the PR information to make a comprehensive analysis.  
The parameter $\beta$ is set to 0.3.
We get $\textbf{F}_{max}$ by adjusting the threshold from 0 to 255.
The $\textbf{F}_{avg}$ is obtained by using twice the mean value of the prediction as the threshold.
Besides, \textbf{Weighted F-measure} ($\textbf{F}^{w}_{\beta}$) , a weighted precision, is designed to improve F-measure.
Due to space constraints, we didn't show $F_{max}$ and $F_{avg}$ in Tab.\ref{T-SOTA1}. 

\subsection{Implementation}
We apply the ResNet-50~\cite{He2016Deep} (pre-trained on ImageNet) as backbone.
All three encoders are ResNet-50 and the 3rd one removes the 1,2 stages of ResNet-50.
Random crop, random horizontal flip, and random size change strategy (288, 320, 352) are used for data augmentation.
Linear decay and warm-up strategies are adopted in the training phase.
Maximum learning rates are set to 5e-3 for the backbone and 5e-2 for other parts. 
Stochastic gradient descent (SGD) is used to train the network (momentum is 0.9, the weight decay is 5e-4). 
Batchsize is set to 16.
The maximum epoch is set to 30.
We train the model on a PC (16GB RAM, RTX 2080Ti GPU).
In the test phase, input images are resized to $352\times352$.

\subsection{Ablation Analysis}

\textbf{Structural analysis}: The first 4 lines of Tab.~\ref{T-XR} are input fusion (IF), early fusion (EF), late result fusion (LF), and three-stream structure (TSS).
The EF and LF can be represented by the brief structure in the lower right corner of Fig.\ref{pipline}.
The ResinRes structure combines the two structures.
Fig.~\ref{early} (d) and the 2nd network in Fig.~\ref{early} (a) show TSS and IF, respectively.
Experiments show that different network structures perform differently on different types of datasets.
In the 6th line (without depth), the FPN structure is used to integrate all levels of features of a single encoder (ResNet-50) without using depth information.
The 7th line is our ResinRes structure without gate unit.
Due to the improvement of network complexity, the overall performance is better than the above structures.
The 8th line indicates that the adaptive-cooperative semantic guidance (ACG) is added to the ResinRes structure.
It is worth noting that here we change the activation function of the features regulated by the gate unit from ReLu to sigmoid to make ACG perform better.
ACG helps the network suppress unreliable information and balance the EF and LF structures, which makes the network more flexible and performs better.
In the last line, type-based attention modules (TAM) are added to further strengthen the network.
The baseline is the same as the network structure of the 6th line experiment (ResNet50+FPN), but only BCE loss is used in the training process, which verifies the effectiveness of IoU loss.
\begin{table*}[htbp]
  \centering
  \caption{
		Quantitative evaluation.
		We compare 18 RGB-D SOD methods on 8 RGB-D datasets.
		The maximum and mean F-measure ($F_{max}$, $F_{avg}$, larger is better), weighted F-measure ($F^{w}_{\beta}$, larger is better), E-measure ($E_{m}$, larger is better), S-measure ($S_{m}$, larger is better) and MAE (smaller is better) of different salient object detection methods is shown in the table.
		The best three results are highlighted in $\textcolor[rgb]{ 1,  0,  0}{red}, \textcolor[rgb]{ .357,  .608,  .835}{blue}$ and $\textcolor[rgb]{ .608,  .733,  .349}{green}$.
		*: traditional methods.
		$e$: a typical algorithm based on edge cues.
		$n$: representative algorithm without depth cues.
		-: the authors of the algorithms do not provide results. 'au' represents ACFNet using data augmentation
		Due to space constraints, we didn't show $F_{max}$ and $F_{avg}$.}

    \resizebox{\textwidth}{!}{
    \begin{tabular}{|l|cccc|cccc|cccc|cccc|}
    \hline
    \multicolumn{1}{|c|}{Model} & \multicolumn{4}{c|}{LFSD Dataset}                                             & \multicolumn{4}{c|}{NJUD Dataset}                                             & \multicolumn{4}{c|}{NLPR Dataset}                                             & \multicolumn{4}{c|}{RGBD135 Dataset} \\
                      & $F^{w}_{\beta}$$\uparrow$                & $E_{m}$$\uparrow$               & $S_{m}$$\uparrow$                & MAE$\downarrow$               & $F^{w}_{\beta}$$\uparrow$               & $E_{m}$$\uparrow$                & $S_{m}$$\uparrow$                & MAE$\downarrow$               & $F^{w}_{\beta}$$\uparrow$                & $E_{m}$$\uparrow$                & $S_{m}$$\uparrow$                & MAE$\downarrow$               & $F^{w}_{\beta}$$\uparrow$                & $E_{m}$                & $S_{m}$$\uparrow$                & MAE$\downarrow$ \\
    \hline
    DES(ICIMCS)$^{*}$    & .272              & .465              & .441              & .414              & .232              & .414              & .415              & .446              & .248              & .733              & .580              & .301              & .292              & .776              & .629              & .291 \\
    DCMC(SPL)$^{*}$      & .579              & .842              & .742              & .161              & .474              & .797              & .689              & .171              & .247              & .684              & .548              & .190              & .159              & .677              & .466              & .191 \\
    CDCP(ICCVW)$^{*}$    & .490              & .737              & .654              & .206              & .479              & .751              & .671              & .188              & .477              & .785              & .731              & .116              & .457              & .810              & .710              & .121 \\
    \hline
    DF(TIP)         & .618              & .841              & .776              & .151              & .521              & .818              & .722              & .158              & .495              & .838              & .763              & .102              & .383              & .806              & .681              & .132 \\
    CTMF(TCYB)      & .695              & .851              & .796              & .120              & .720              & .866              & .849              & .085              & .679              & .869              & .860              & .056              & .686              & .911              & .863              & .055 \\
    PCANet(CVPR)    & .714              & .846              & .800              & .113              & .803              & .909              & .877              & .059              & .762              & .916              & .874              & .044              & .710              & .912              & .843              & .050 \\
    AFNet(ACCESS)   & .671              & .810              & .738              & .133              & .696              & .847              & .772              & .100              & .693              & .884              & .799              & .058              & .641              & .874              & .770              & .068 \\
    CPFP(CVPR)      & .775              & .867              & .828              & .088              & .828              & .900              & .878              & .053              & .813              & .924              & .888              & .036              & .787              & .927              & .872              & .038 \\
    MMCI(PR,IROS) & .663              & .840              & .787              & .132              & .739              & .882              & .859              & .079              & .676              & .872              & .856              & .059              & .650              & .904              & .848              & .065 \\
    TANet(TIP)      & .718              & .845              & .801              & .111              & .803              & .893              & .878              & .061              & .778              & .916              & .886              & .041              & .739              & .919              & .858              & .046 \\
    DMRA(ICCV)      & \textcolor[rgb]{ 1,  0,  0}{.811} & \textcolor[rgb]{ 1,  0,  0}{.899} & \textcolor[rgb]{ .267,  .447,  .769}{.847} & \textcolor[rgb]{ 1,  0,  0}{.075} & .847              & .919              & .886              & .051              & .839              & .942              & .899              & .031              & .843              & .944              & .900              & .030 \\
    D3Net(TNNLS)    & .756              & .860              & .832              & .099              & .833              & .901              & .895              & .051              & .826              & .934              & .906              & .034              & .831              & .956              & .904              & .030 \\
    S2MA(CVPR)      & .772              & .876              & .837              & .094              & .842              & .916              & .894              & .053              & .852              & .942              & .915              & .030              & \textcolor[rgb]{ 1,  0,  0}{.892} & \textcolor[rgb]{ 1,  0,  0}{.974} & \textcolor[rgb]{ 1,  0,  0}{.941} & \textcolor[rgb]{ 1,  0,  0}{.021} \\
    DANet(ECCV)     & \textcolor[rgb]{ .439,  .678,  .278}{.789} & .878 & \textcolor[rgb]{ .439,  .678,  .278}{.845} & \textcolor[rgb]{ .439,  .678,  .278}{.082} & \textcolor[rgb]{ .439,  .678,  .278}{.853} & \textcolor[rgb]{ .267,  .447,  .769}{.926} & .897              & .046              & \textcolor[rgb]{ .439,  .678,  .278}{.858} & \textcolor[rgb]{ .439,  .678,  .278}{.949} & .915              & .028              & \textcolor[rgb]{ .439,  .678,  .278}{.848} & \textcolor[rgb]{ .439,  .678,  .278}{.961} & .905              & .028 \\
    DCF(CVPR)       & .788              & \textcolor[rgb]{ .267,  .447,  .769}{.884} & .841              & .082              & \textcolor[rgb]{ .267,  .447,  .769}{.868} & \textcolor[rgb]{.439,  .678,  .278}{.924} & \textcolor[rgb]{ .267,  .447,  .769}{.911} & \textcolor[rgb]{ .267,  .447,  .769}{.044} & \textcolor[rgb]{ .267,  .447,  .769}{.867} & \textcolor[rgb]{ 1,  0,  0}{.957} & \textcolor[rgb]{ .267,  .447,  .769}{.923} & \textcolor[rgb]{ .267,  .447,  .769}{.026} & .840              & .950              & .904              & .028 \\
    cmSalGAN(TMM)   & .761              & .877              & .830              & .097              & .846              &  .923  & \textcolor[rgb]{ .439,  .678,  .278}{.903} & \textcolor[rgb]{ .439,  .678,  .278}{.046} & .855              & \textcolor[rgb]{ .439,  .678,  .278}{.949} & \textcolor[rgb]{ .439,  .678,  .278}{.922} & \textcolor[rgb]{ .439,  .678,  .278}{.027} & .840              & .948              & \textcolor[rgb]{ .439,  .678,  .278}{.913} & \textcolor[rgb]{ .439,  .678,  .278}{.028} \\
    \hline
    SCRN(ICCV)$^{e}$     & .728              & .862              & .813              & .109              & .840              & .922              & .899              & .047              & .833              & .940              & .912              & .032              & .809              & .939              & .896              & .033 \\
    \hline
    GCPANet(AAAI)$^{n}$  & .745              & .870              & .822              & .105              & .844              & .922              & .901              & .049              & .854              & .944              & .919              & .028              & .836              & .953              & .908              & .031 \\
    \hline
    ACFNet(au)        & \textcolor[rgb]{ .267,  .447,  .769}{.802}              & \textcolor[rgb]{ .439,  .678,  .278}{.882}              & \textcolor[rgb]{ 1,  0,  0}{.852}              & \textcolor[rgb]{ .267,  .447,  .769}{.079}              & \textcolor[rgb]{ 1,  0,  0}{.883} & \textcolor[rgb]{ 1,  0,  0}{.936} & \textcolor[rgb]{ 1,  0,  0}{.914} & \textcolor[rgb]{ 1,  0,  0}{.037}
    & \textcolor[rgb]{ 1,  0,  0}{.881} & \textcolor[rgb]{ .267,  .447,  .769}{.955} & \textcolor[rgb]{ 1,  0,  0}{.924} & \textcolor[rgb]{ 1,  0,  0}{.025} & \textcolor[rgb]{ .267,  .447,  .769}{.873} & \textcolor[rgb]{ .267,  .447,  .769}{.965}              & \textcolor[rgb]{ .267,  .447,  .769}{.915}              & \textcolor[rgb]{ .267,  .447,  .769}{.022} \\
    \hline  
    \hline
    \multicolumn{1}{|c|}{Model} & \multicolumn{4}{c|}{SIP Dataset}                                              & \multicolumn{4}{c|}{SSD Dataset}                                              & \multicolumn{4}{c|}{STEREO Dataset}                                           & \multicolumn{4}{c|}{DUTRGBD Dataset} \\
                      & $F^{w}_{\beta}$ $\uparrow$                & $E_{m}$ $\uparrow$                  & $S_{m}$   $\uparrow$                & MAE$\downarrow$               & $F^{w}_{\beta}$ $\uparrow$                & $E_{m}$ $\uparrow$                  & $S_{m}$   $\uparrow$                 & MAE$\downarrow$               & $F^{w}_{\beta}$ $\uparrow$                & $E_{m}$ $\uparrow$                  & $S_{m}$   $\uparrow$                 & MAE$\downarrow$               & $F^{w}_{\beta}$ $\uparrow$               & $E_{m}$ $\uparrow$                 & $S_{m}$   $\uparrow$                & MAE$\downarrow$ \\
    \hline
    DES(ICIMCS)$^{*}$    & .338              & .746              & .613              & .298              & .171              & .379              & .343              & .497              & .370              & .687              & .639              & .295              & .372              & .726              & .657              & .281 \\
    DCMC(SPL)$^{*}$      & .413              & .786              & .683              & .186              & .458              & .789              & .695              & .171              & .520              & .831              & .731              & .148              & .269              & .713              & .494              & .240 \\
    CDCP(ICCVW)$^{*}$    & .397              & .722              & .595              & .224              & .403              & .713              & .608              & .219              & .558              & .796              & .713              & .149              & .491              & .794              & .685              & .165 \\
    \hline
    DF(TIP)         & .406              & .794              & .653              & .185              & .510              & .801              & .731              & .157              & .549              & .838              & .757              & .141              & .514              & .842              & .719              & .150 \\
    CTMF(TCYB)      & .535              & .824              & .716              & .139              & .621              & .837              & .776              & .100              & .698              & .870              & .848              & .086              & .681              & .882              & .830              & .097 \\
    PCANet(CVPR)    & .768              & .900              & .842              & .071              & .730              & .884              & .843              & .064              & .778              & .907              & .875              & .064              & .683              & .858              & .800              & .101 \\
    AFNet(ACCESS)   & .617              & .815              & .720              & .118              & .589              & .803              & .714              & .118              & .752              & .887              & .825              & .075              & -                 & -                 & -                 & - \\
    CPFP(CVPR)      & .788              & .899              & .850              & .064              & .708              & .832              & .807              & .082              & .817              & .907              & .879              & .051              & .636              & .815              & .749              & .100 \\
    MMCI(PR,IROS) & .712              & .886              & .833              & .086              & .660              & .860              & .814              & .082              & .760              & .905              & .873              & .068              & .626              & .855              & .791              & .113 \\
    TANet(TIP)      & .748              & .894              & .835              & .075              & .726              & .879              & .839              & .063              & .787              & .916              & .871              & .060              & .703              & .866              & .808              & .093 \\
    DMRA(ICCV)      & .734              & .858              & .800              & .088              & .786              & .896              & .857              & .058              & .647              & .816              & .752              & .086              & .852              & .930              & \textcolor[rgb]{ .439,  .678,  .278}{.888} & .048 \\
    D3Net(TNNLS)    & .793              & .903              & .864              & .063              & .780              & .892              & .866              & .058              & .815              & .911              & .891              & .054              & -                 & -                 & -                 & - \\
    S2MA(CVPR)      & -                 & -                 & -                 & -                 & .787              & .898              & \textcolor[rgb]{ .439,  .678,  .278}{.868} & \textcolor[rgb]{ .439,  .678,  .278}{.052} & .825              & \textcolor[rgb]{ .439,  .678,  .278}{.926} & .890              & .051              & \textcolor[rgb]{ .439,  .678,  .278}{.856} & .921              & \textcolor[rgb]{ .267,  .447,  .769}{.903} & .046 \\
    DANet(ECCV)     & \textcolor[rgb]{ .267,  .447,  .769}{.829} & \textcolor[rgb]{ .439,  .678,  .278}{.917} & \textcolor[rgb]{ .267,  .447,  .769}{.878} & \textcolor[rgb]{ .267,  .447,  .769}{.054} & \textcolor[rgb]{ .267,  .447,  .769}{.798} & \textcolor[rgb]{ .267,  .447,  .769}{.909} & \textcolor[rgb]{ .267,  .447,  .769}{.869} & \textcolor[rgb]{ .267,  .447,  .769}{.050} & .830              & \textcolor[rgb]{ .439,  .678,  .278}{.926} & .892              & .047              & -                 & -                 & -                 & - \\
    DCF(CVPR)       & \textcolor[rgb]{ .439,  .678,  .278}{.819} & \textcolor[rgb]{ .267,  .447,  .769}{.920} & \textcolor[rgb]{ .439,  .678,  .278}{.874} & .059              & \textcolor[rgb]{ .439,  .678,  .278}{.787} & .898              & .864              & .057              & \textcolor[rgb]{ .267,  .447,  .769}{.851} & \textcolor[rgb]{.267,  .447,  .769}{.929} & \textcolor[rgb]{ .267,  .447,  .769}{.902} & \textcolor[rgb]{ .267,  .447,  .769}{.045} & -              & -              & -              & - \\
    cmSalGAN(TMM)   & .795              & .905              & .865              & .064              & .650              & .851              & .791              & .086              & -                 & -                 & -                 & -                 & .786              & .897              & .867              & .068 \\
    \hline
    SCRN(ICCV)$^{e}$     & .803              & .914              & .871              & \textcolor[rgb]{ .439,  .678,  .278}{.058} & .774              & \textcolor[rgb]{ .439,  .678,  .278}{.902} & .864              & .054              & \textcolor[rgb]{ .439,  .678,  .278}{.833} & \textcolor[rgb]{ .267,  .447,  .769}{.929} & \textcolor[rgb]{ .439,  .678,  .278}{.901} & \textcolor[rgb]{ .439,  .678,  .278}{.046} & .856              & \textcolor[rgb]{ .267,  .447,  .769}{.942} & \textcolor[rgb]{ 1,  0,  0}{.909} & \textcolor[rgb]{ .439,  .678,  .278}{.043} \\
    \hline
    GCPANet(AAAI)$^{n}$  & .773              & .897              & .854              & .073              & .731              & .870              & .841              & .067              & .827              & .921              & .897              & .052              & \textcolor[rgb]{ .267,  .447,  .769}{.863} & \textcolor[rgb]{ .267,  .447,  .769}{.942} & \textcolor[rgb]{ 1,  0,  0}{.909} & \textcolor[rgb]{ .267,  .447,  .769}{.042} \\
    \hline
    ACFNet(au)        & \textcolor[rgb]{ 1,  0,  0}{.847} & \textcolor[rgb]{ 1,  0,  0}{.927} & \textcolor[rgb]{ 1,  0,  0}{.887} & \textcolor[rgb]{ 1,  0,  0}{.046} & \textcolor[rgb]{ 1,  0,  0}{.805} & \textcolor[rgb]{ 1,  0,  0}{.912} & \textcolor[rgb]{ 1,  0,  0}{.871} & \textcolor[rgb]{ 1,  0,  0}{.046} & \textcolor[rgb]{ 1,  0,  0}{.859} & \textcolor[rgb]{ 1,  0,  0}{.935} & \textcolor[rgb]{ 1,  0,  0}{.904} & \textcolor[rgb]{ 1,  0,  0}{.040} & \textcolor[rgb]{ 1,  0,  0}{.871} & \textcolor[rgb]{ 1,  0,  0}{.943} & \textcolor[rgb]{ 1,  0,  0}{.909} & \textcolor[rgb]{ 1,  0,  0}{.041} \\

    \hline
    \end{tabular}%
    }
  \label{T-SOTA1}%
\end{table*}%

\begin{table*}[htbp]
	\centering
	\caption{Ablation experiments.
		The 1st, 2nd 3rd and 4th lines are input fusion (IF), early fusion (EF), late result fusion (LF), and three-stream structure (TSS).
		The 5th line (w/o depth) is obtained by using only Decoder1-R and FPN.
		Depth images are not used.
		ResinRes is the network we designed.
		The 7th line uses the ACG. 
		The 8th line adds TAM based on the 7th line. 
		We follow the previous work~\cite{2020-TIP-DPANet} to use part of the datasets (relatively large datasets) to do ablation experiments.
	}
	\resizebox{\textwidth}{!}{
		\begin{tabular}{|l|cccc|cccc|cccc|cccc|}
			\hline
			\multicolumn{1}{|c|}{Model} & \multicolumn{4}{c|}{STEREO Dateset}                                                   & \multicolumn{4}{c|}{SIP Dateset}                                                      & \multicolumn{4}{c|}{NJUD Dateset}                                                     & \multicolumn{4}{c|}{LFSD Dateset} \\
			&  $F^{w}_{\beta}$ $\uparrow$                & $E_{m}$$\uparrow$                & $S_{m}$$\uparrow$                & MAE $\downarrow$              & $F^{w}_{\beta}$ $\uparrow$                & $E_{m}$ $\uparrow$               & $S_{m}$$\uparrow$                & MAE$\downarrow$                &  $F^{w}_{\beta}$$\uparrow$                & $E_{m}$ $\uparrow$               & $S_{m}$ $\uparrow$               & MAE $\downarrow$               & $F^{w}_{\beta}$ $\uparrow$                & $E_{m}$$\uparrow$                & $S_{m}$$\uparrow$                & MAE$\downarrow$  \\
			\hline
			IF                & .831              & .914              & .892              & .049              & .791              & .902              & .863              & .065              & .846              & .910              & .896              & .050              & .790              & .874 & .847              & .081 \\
			EF                & .838              & .915              & .896              & .048              & .820              & .918              & .879              & .055              & .851              & .914              & .900              & .050              & .773              & .859              & .837              & .091 \\
			LF                & .845              & .918              & .902              & .045              & .824              & .917              & .884              & .053              & .863              & .909              & .909              & .045              & .752              & .847              & .828              & .094 \\
			TSS               & .839              & .920              & .899              & .046              & .817              & .920              & .881              & .055              & .856              & .912              & .905              & .048              & .787              & .859              & .843              & .089 \\
			\hline
			Baseline         & . 820              & .911               & .895               & .050               & .792               & .909               & .865              & .062              & .822              & .896             & .890             & .053              & .696              & .826             & .790              & .121 \\
			w/o depth         & .843              & .922              & .902              & .046              & .813              & .919              & .873              & .058              & .849              & .916              & .900              & .048              & .727              & .820              & .802              & .112 \\
			ResinRes          & .850              & .918              & .906              & .044              & .821              & .921              & .883              & .055              & .859              & .910              & .909              & .046              & .776              & .853              & .837              & .091 \\
			ResinRes+G        & \textcolor[rgb]{ 1,  0,  0}{.860} & .920              & \textcolor[rgb]{ 1,  0,  0}{.907} & .041              & .829              & .918              & .880              & .051              & .873              & .910              & .912              & .040              & .784              & .855              & .837              & .089 \\
			ResinRes+G+TAM    & .856              & \textcolor[rgb]{ 1,  0,  0}{.922} & .904              & \textcolor[rgb]{ 1,  0,  0}{.041} & \textcolor[rgb]{ 1,  0,  0}{.841} & \textcolor[rgb]{ 1,  0,  0}{.925} & \textcolor[rgb]{ 1,  0,  0}{.887} & \textcolor[rgb]{ 1,  0,  0}{.048} & \textcolor[rgb]{ 1,  0,  0}{.876} & \textcolor[rgb]{ 1,  0,  0}{.919} & \textcolor[rgb]{ 1,  0,  0}{.914} & \textcolor[rgb]{ 1,  0,  0}{.039} & \textcolor[rgb]{ 1,  0,  0}{.800} & \textcolor[rgb]{ 1,  0,  0}{.869}              & \textcolor[rgb]{ 1,  0,  0}{.853} & \textcolor[rgb]{ 1,  0,  0}{.080} \\
			\hline
		\end{tabular}%
	}
	\label{T-XR}%
\end{table*}%

\begin{table*}[htbp]
	\centering
	\caption{Analysis of type-based attention module (TAM).
		The 1st line is the results of ACFNet (Fig.\ref{pipline}), where TAM is shown in Fig.\ref{TAM}.
		In the 3rd line, the MLP of each TAM is removed and the weights of features obtained by various types of convolution are not recalibrated.
		The 2nd line (CAM-SE) adds channel attention modules after concatenation operation on the basis of the 3rd line.
		The channel attention mechanism we use here is SE block~\cite{2018-CVPR-SENet}.
	}
	\resizebox{\textwidth}{!}{
		\begin{tabular}{|c|cccc|cccc|cccc|cccc|}
			\hline
			Model             & \multicolumn{4}{c|}{STEREO}                                                   & \multicolumn{4}{c|}{SIP}                                                      & \multicolumn{4}{c|}{NJUD}                                                     & \multicolumn{4}{c|}{LFSD} \\
			&  $F^{w}_{\beta}$ $\uparrow$                & $E_{m}$$\uparrow$                & $S_{m}$$\uparrow$                & MAE $\downarrow$              & $F^{w}_{\beta}$ $\uparrow$                & $E_{m}$ $\uparrow$               & $S_{m}$$\uparrow$                & MAE$\downarrow$                &  $F^{w}_{\beta}$$\uparrow$                & $E_{m}$ $\uparrow$               & $S_{m}$ $\uparrow$               & MAE $\downarrow$               & $F^{w}_{\beta}$ $\uparrow$                & $E_{m}$$\uparrow$                & $S_{m}$$\uparrow$                & MAE$\downarrow$  \\
			\hline
			TAM               & \textcolor[rgb]{ 1,  0,  0}{.856} & \textcolor[rgb]{ 1,  0,  0}{.922} & \textcolor[rgb]{ 1,  0,  0}{.904} & \textcolor[rgb]{ 1,  0,  0}{.041} & \textcolor[rgb]{ 1,  0,  0}{.841} & \textcolor[rgb]{ 1,  0,  0}{.925} & \textcolor[rgb]{ 1,  0,  0}{.887} & \textcolor[rgb]{ 1,  0,  0}{.048} & \textcolor[rgb]{ 1,  0,  0}{.876} & \textcolor[rgb]{ 1,  0,  0}{.919} & \textcolor[rgb]{ 1,  0,  0}{.914} & \textcolor[rgb]{ 1,  0,  0}{.039} & \textcolor[rgb]{ 1,  0,  0}{.800} & \textcolor[rgb]{ 1,  0,  0}{.869} & \textcolor[rgb]{ 1,  0,  0}{.853} & \textcolor[rgb]{ 1,  0,  0}{.080} \\
			CAM-SE            & .855              & .917              & .902              & \textcolor[rgb]{ 1,  0,  0}{.041} & .826              & .916              & .876              & .052              & \textcolor[rgb]{ 1,  0,  0}{.878} & .908              & \textcolor[rgb]{ 1,  0,  0}{.914} & \textcolor[rgb]{ 1,  0,  0}{.039} & .780              & .859              & .836              & .087 \\
			TAM-w/o MLP       & .851              & .918              & .903              & .042              & .834              & .923              & .886              & .049              & .874              & .915              & .914              & .040              & .786              & .867              & .841              & .086 \\
			\hline
		\end{tabular}%
	}
	\label{T-TAM}%
\end{table*}%

\begin{figure*}[htb]%
	\centering
	\includegraphics[width=2\columnwidth, height=17cm]{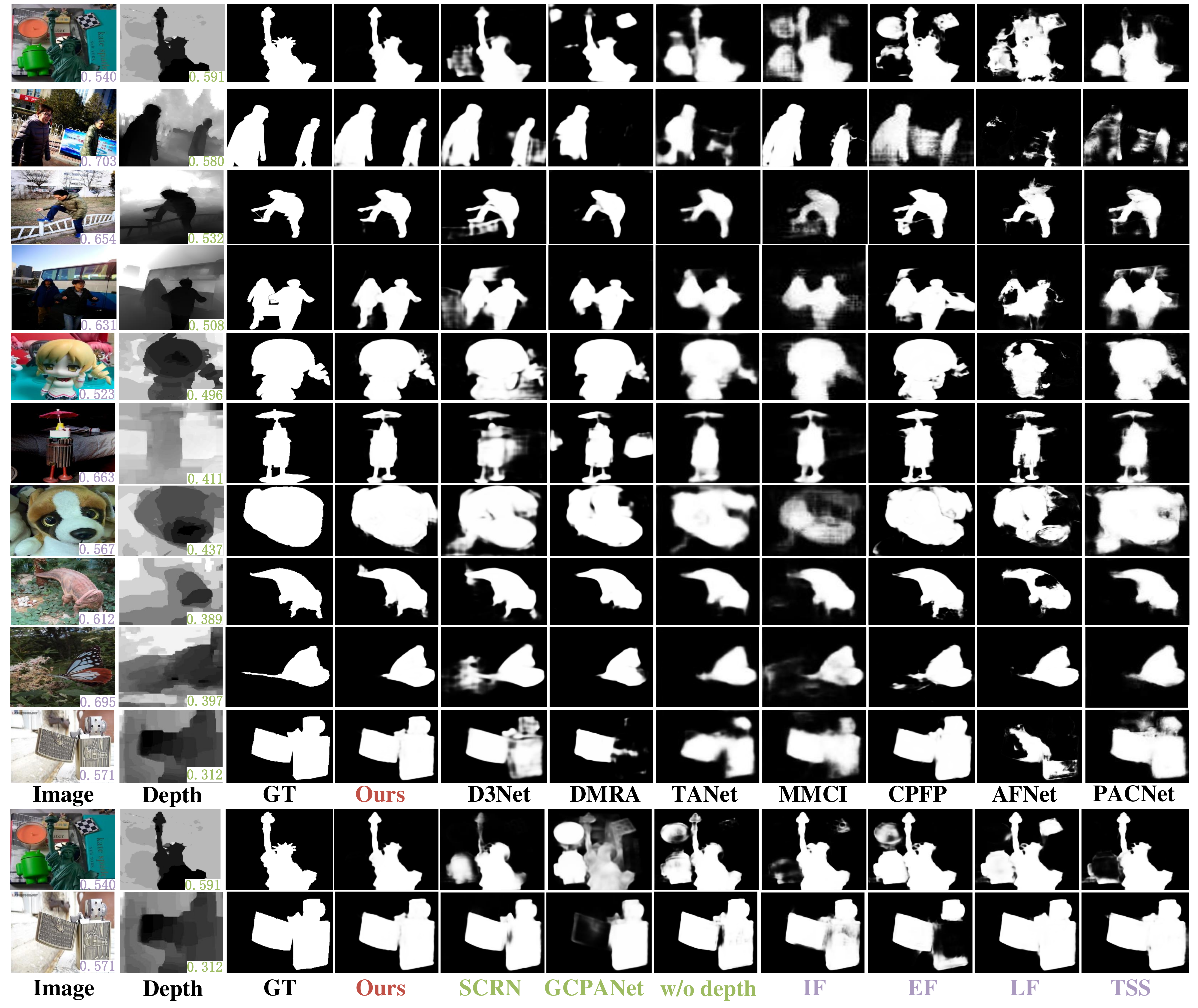}
	\caption{Qualitative comparisons with state-of-the-art algorithms.
		From top to bottom, the credibility and utilization value of depth images are decreasing.
		In the last two rows, the results of the algorithms (SCRN, GCPANet, ResNet-50+FPN) without depth data and the saliency maps of different structures (IF, EF, LF, TSS) are compared.
	}
	\label{compare}
\end{figure*}

\textbf{Module analysis}: To further verify the effectiveness of the proposed TAM module, more comparative experiments are shown in Tab.\ref{T-TAM}.
The 1st line is the final network ACFNet with TAMs, as shown in the Fig.\ref{pipline}.
The 3rd line (TAM w/o MLP) removes the multi-layer perceptron (MLP) from the TAM.
MLP is the key to adaptively adjust the influence of different convolution features.
After removing the weight evaluation (MLP) of each type of feature in TAM, the effect of TAM becomes worse, as shown in Tab.~\ref{T-TAM} 3rd line.
The 2nd line (CAM-SE) adds a channel attention module (SE block~\cite{2018-CVPR-SENet}) after concatenation operation on the basis of the 3rd line.
It is worth noting that our type-based attention is to evaluate the weight of a group of features obtained by convolutions of the same type and balance the features by weighting.
CAM-SE directly adds different weights to all features in the channel dimension.
CAM-SE is the traditional channel attention, while our method focuses more on the adaptive cooperation of different types of convolutions.
Experiments show that TAM performs best.

\textbf{Decoder analysis}:
The basic structure of the decoder can be divided into 2 types (parallel and progressive).
FPN is representative of progressive.
Hypercolumn~\cite{2015-CVPR-Hypercolumn} is a representative of parallel.
We make many attempts on the decoder structure of Encoder1-R and Encoder1-D respectively.
As shown in Tab.\ref{FP}, R and D represent the decoder structure of Encoder1-R and Encoder1-D respectively.
F represents the FPN structure.
H represents the Hypercolumn structure.
After experiments, our network decoders use the FPN structure uniformly.

\begin{table}[htbp]
  \centering
  \scriptsize
  \caption{Decoder analysis. R and D represent the decoder structure of Encoder1-R and Encoder1-D respectively. F represents the FPN structure.
H represents the Hypercolumn structure.}
  \setlength{\tabcolsep}{0.7mm}{
    \begin{tabular}{|l|rr|rr|rr|rr|}
    \hline
                      & \multicolumn{2}{c|}{STEREO}           & \multicolumn{2}{c|}{NJUD }           & \multicolumn{2}{c|}{SIP}              & \multicolumn{2}{c|}{LFSD} \\
    Model             & \multicolumn{1}{l}{$F^{w}_{\beta}$ $\uparrow$} & \multicolumn{1}{l|}{MAE$\downarrow$} & \multicolumn{1}{l}{$F^{w}_{\beta}$ $\uparrow$} & \multicolumn{1}{l|}{MAE$\downarrow$} & \multicolumn{1}{l}{$F^{w}_{\beta}$ $\uparrow$} & \multicolumn{1}{l|}{MAE$\downarrow$} & \multicolumn{1}{l}{$F^{w}_{\beta}$ $\uparrow$} & \multicolumn{1}{l|}{MAE$\downarrow$} \\
    \hline
    D-F, R-H       & .855              & .041              & \textcolor[rgb]{ 1,  0,  0}{.883} & \textcolor[rgb]{ 1,  0,  0}{.037} & .830              & .051              & .781              & .092 \\
    D-H, R-F       & .858              & .041              & .879              & .038              & .835              & .050              & .760              & .094 \\
    D-H, R-H      & .855              & .041              & \textcolor[rgb]{ 1,  0,  0}{.883} & .038              & .842              & .047              & .779              & .087 \\
    D-F, R-F       & \textcolor[rgb]{ 1,  0,  0}{.859} & \textcolor[rgb]{ 1,  0,  0}{.040} & \textcolor[rgb]{ 1,  0,  0}{.883} & \textcolor[rgb]{ 1,  0,  0}{.037} & \textcolor[rgb]{ 1,  0,  0}{.847} & \textcolor[rgb]{ 1,  0,  0}{.046} & \textcolor[rgb]{ 1,  0,  0}{.802} & \textcolor[rgb]{ 1,  0,  0}{.079} \\
    \hline
    \end{tabular}%
    }
  \label{FP}%
\end{table}%

\subsection{Analysis of ACG}
\begin{figure*}[htp]
	\centering
	\includegraphics[width=2\columnwidth]{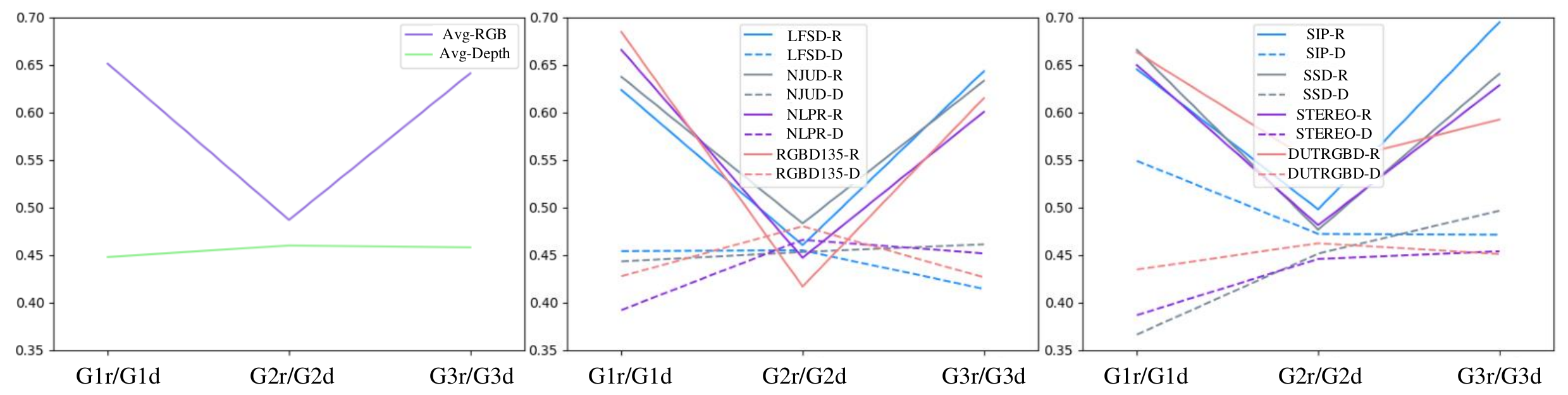} 
	\caption{Analysis of adaptively-cooperative semantic guidance.
		The left figure shows the average weights of $G1_{r}$, $G2_{r}$, $G3_{r}$, $G1_{d}$, $G2_{d}$, $G3_{d}$ positions in all test datasets.
		The green line and purple line represent the weights of depth features and RGB features, respectively.
		The 2nd and 3rd figures show the independent analysis of 8 test datasets. 
	}
	\label{gate}
\end{figure*}

We show the ACG data analysis in Fig.~\ref{gate}.
The left figure of Fig.~\ref{gate} shows the average weights of $G1_{r}$, $G2_{r}$, $G3_{r}$, $G1_{d}$, $G2_{d}$, $G3_{d}$ positions in all test datasets.
The green line and purple line represent the weights of depth features and RGB features, respectively.
The 2nd and 3rd figures in Fig.~\ref{gate} show the independent analysis of 8 test datasets. 
Here we use solid lines to represent $G1_{r}$, $G2_{r}$, $G3_{r}$ and dotted lines to represent $G1_{d}$, $G2_{d}$, $G3_{d}$.
We can find that depth features are generally suppressed, which indicates that RGB data plays a leading role.
In some datasets, depth information is suppressed seriously because of its low reliability.
Because the depth information of RGBD135 is relatively reliable, we can even see that $G2_{d}>G2_{r}$.
Besides, because feature $R4'$ is the deepest RGB feature with rich semantic information, its weight is usually high.
$R$ is the final output feature of pure RGB data through the FPN structure, so its weight is also high.
The above regulation is a global dynamic process.

We show the values of $G1_{r}$ and $G1_{d}$ for specific images in Fig.~\ref{compare}.
The green number and the purple number represent the values of the weight $G1_{r}$ and $G1_{d}$ of the RGB and Depth data of each picture respectively.
From top to bottom, the credibility and utilization value of depth images are declining.
This is consistent with our intuitive feelings.

\begin{figure*}[htb]%
	\centering
	\includegraphics[width=1.7\columnwidth, height=15cm]{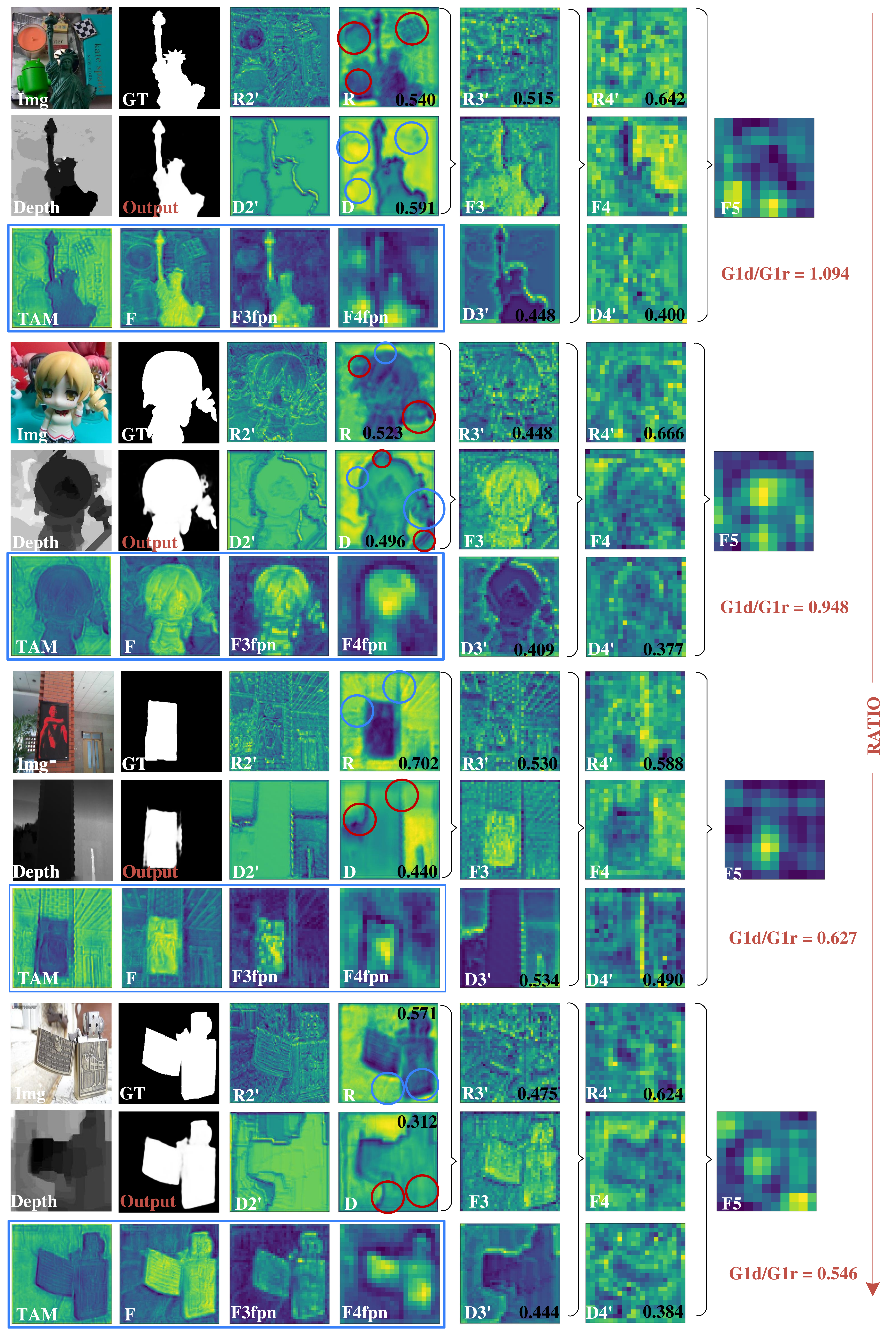}
	\caption{Visualizations of features.
		The $\textcolor[rgb]{ 1,  0,  0}{red}$  and $\textcolor[rgb]{ .357,  .608,  .835}{blue}$ \textbf{circles} represent the wrong and correct positions of the single-mode data analysis results, respectively.
		The visualization results of the features of each position in decoder2 are shown in the $\textcolor[rgb]{ .357,  .608,  .835}{blue}$ \textbf{boxes}.
		The black \textbf{arrows} indicate the feature transfer process of encoder2.
		From top to bottom, the ratio (G1$_{d}$/G1$_{r}$) is getting smaller. ACG makes the network more believe in the guidance of RGB feature ($R$) and suppresses the depth feature ($D$).
	}
	\label{tosee}
\end{figure*}

On the penultimate row (Fig.~\ref{compare}), our model spontaneously find that the depth feature has high credibility, so it automatically gives a high weight (G1$_{d}$ = 0.591) to the depth feature of the $\textit{statue of liberty}$, which helps our model detect more accurately.
Compared with the algorithms without depth information (SCRN~\cite{2019-ICCV-SCRN}, GCPANet~\cite{2020-AAAI-GCPANet}, ResNet50+FPN) and other structures (IF, EF, LF, TSS), our algorithm has obvious advantages.
In the last row, GCPANet misjudged the left half of the $\textit{lighter}$ because it didn't use the depth data.
In addition, the EF structure in the last row and the DMRA algorithm~\cite{2019-ICCV-DMRA} in the 10th row both rely too much on the depth information (the right half of the depth data is inaccurate), resulting in misjudgment of the right half of the $\textit{lighter}$.
The above experiments show that ACG plays a key role in balancing the influence of different types of features.

\subsection{Visual Analysis of The Second Phase}
We show the feature visualization results of Encoder2 and Decoder2 in Fig.\ref{tosee}.
In Encoder2, $F3$ is obtained from the low-level features $R2'$, $D2'$ and guidance features $R, D$.
$F4$ is obtained from features $R3'$, $D3'$ and $F3$.
$F5$ is obtained from features $R4'$, $D4'$ and $F4$.
The above inferring process is consistent with Fig.\ref{pipline}.
We add red circles to the features $R$ and $D$ obtained from the single-mode data to indicate the error locations.
The blue circles highlight the correct part of the features.
The weights of the features regulated by ACG are marked by black numbers.
The features in the blue boxes are the feature maps of each stage of the Decoder2 (FPN), which are consistent with Eq.6-8.
Feature TAM is the output feature optimized by the TAM module.

In Fig.\ref{tosee}, the depth maps of the first two examples are relatively accurate, so the weight of feature $D$ is relatively high.
The feature $D$ in the first example plays a leading role due to its accuracy and successfully suppresses the background interference in feature $R$.
The features $R$ and $D$ in the second picture both have inaccurate positions, and these noises are successfully suppressed by adaptive cooperative fusion (ACG).
The depth features in the last two examples will cause false guidance, so the weights are very low.
The depth data in the third example is not helpful for the specific location of the object, so the weight of $R$ is far greater than that of $D$.
All the above processes are realized spontaneously by the network, which can prove the effectiveness of adaptively cooperative semantic guidance.

\subsection{Comparision with State-of-the arts}

\textbf{RGB-D salient object detection:} We compare the proposed method with 18 state-of-the-art methods, including DES~\cite{2014-ICIMCS-DES}, DCMC~\cite{2016-SPL-DCMC}, CDCP~\cite{2017-ICCVW-CDCP}, DF~\cite{2017-TIP-DF}, CTMF~\cite{2017-TCYBER-CTMF}, PCANet~\cite{2018-CVPR-PCANet}, AFNet~\cite{2019-ACCESS-AFNet}, CPFP~\cite{2019-CVPR-CPFP}, MMCI~\cite{2018-PR-MMCI}, TANet~\cite{2019-TIP-TANet}, DMRA~\cite{2019-ICCV-DMRA}, D3Net~\cite{2020-TNNLS-D3Net}, S2MA~\cite{2020-CVPR-S2MA}, DANet~\cite{2020-ECCV-DANet}, DCF~\cite{2021-CVPR-DCF}, cmSalGAN~\cite{2021-TMM-cmSalGAN}, SCRN~\cite{Wu2019ICCV}, GCPANet~\cite{2020-AAAI-GCPANet}.  
For fair comparisons, we use the saliency maps provided by the authors or generated by the codes they provided.

\textbf{Multi-spectral salient object detection:} To verify the universality of our model, we also verify our model on the new multi-spectral salient object detection (RGB-NIR) dataset~\cite{2020-TMM-RGBT}, as shown in Tab.\ref{NIR}.
The dataset contains 780 pairs of images.
Each pair of images contains RGB images and near-infrared (NIR) images.
We follow~\cite{2020-TMM-RGBT} to complete the experiments of supervised learning and unsupervised learning.
In supervised learning, we complete the training in the training dataset of RGB-NIR SOD dataset and the test in the test dataset of RGB-NIR SOD dataset.
In unsupervised learning, we adopt the same scheme to generate pseudo-NIR images of training dataset (MSRA-B) as~\cite{2020-TMM-RGBT}.
We use MSRA-B (with paired pseudo-NIR images) as a training dataset.
We test the trained model in the whole RGB-NIR dataset (780).
Here we just verify the universality of the ACFNet. For details of experiment, please refer to~\cite{2020-TMM-RGBT}.

We compare the proposed method with 6 state-of-the-art methods, including RC~\cite{2015-TPAMI-GC}, LRK\cite{2012-CVPR-LRK}, CWS~\cite{2013-TIP-CWS}, FT~\cite{2009-CVPR-FT}, DCL~\cite{2016-CVPR-DCL}, SOD8s+~\cite{2020-TMM-RGBT}.
Experiments show that our model also performs best in multi-spectral salient object detection task.

\begin{table}[htbp]
  \centering
  \scriptsize
  \caption{Quantitative evaluation for RGB-NIR dataset. We compare 6 methods on RGB-NIR dataset.
		The maximum and mean F-measure ($F_{max}$, $F_{avg}$, larger is better), MAE (smaller is better) of different salient object detection methods is shown in the table.
		The best results are highlighted in $\textcolor[rgb]{ 1,  0,  0}{red}$.}
    \begin{tabular}{|l|rrr|}
    \hline
    \multicolumn{4}{|c|}{unsupervised} \\
    \hline
    Model             & \multicolumn{1}{l}{$F_{max}$$\uparrow$} & \multicolumn{1}{l}{$F_{avg}$$\uparrow$} & \multicolumn{1}{l|}{MAE$\downarrow$} \\
    \hline
    RC                & .733              & .664              & .146 \\
    LRK               & .664              & .566              & .179 \\
    CWS               & .578              & .506              & .245 \\
    FT                & .042              & .370              & .193 \\
    DCL               & .837              & .779              & .077 \\
    SOD8s+            & .850              & .803              & .061 \\
    \hline
    ACFNet            & \textcolor[rgb]{ 1,  0,  0}{.895} & \textcolor[rgb]{ 1,  0,  0}{.859} & \textcolor[rgb]{ 1,  0,  0}{.035} \\
    \hline
    \multicolumn{1}{r}{} &                   &                   & \multicolumn{1}{r}{} \\
    \hline
    \multicolumn{4}{|c|}{supervised} \\
    \hline
    Model             & \multicolumn{1}{l}{$F_{max}$$\uparrow$} & \multicolumn{1}{l}{$F_{avg}$$\uparrow$} & \multicolumn{1}{l|}{MAE$\downarrow$} \\
    \hline
    SOD8s+            & .903              & .853              & .039 \\
    \hline
    ACFNet            & \textcolor[rgb]{ 1,  0,  0}{.924} & \textcolor[rgb]{ 1,  0,  0}{.867} & \textcolor[rgb]{ 1,  0,  0}{.032} \\
    \hline
    \end{tabular}%
  \label{NIR}%
\end{table}%

\subsection{Quantitative Evaluation}
\begin{figure*}
	
	\label{PAAbefore}
	\centering
	 {
		\begin{minipage}{15cm}
                        \includegraphics[width=\textwidth]{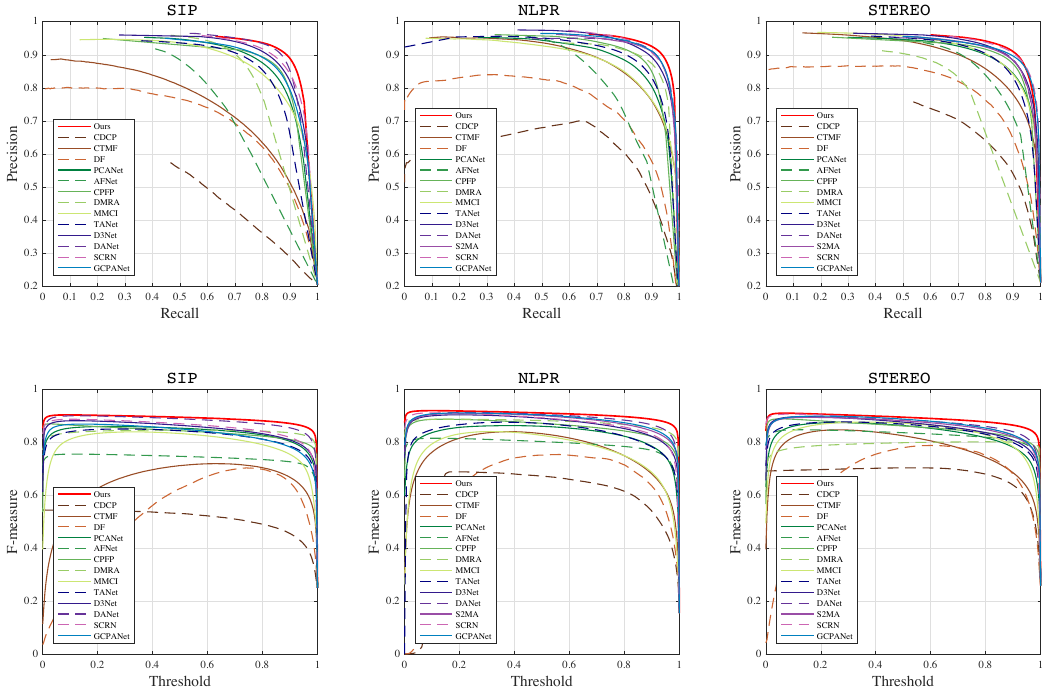} \\
		\end{minipage}
	}
	{
		\begin{minipage}{15cm}
			\includegraphics[width=\textwidth]{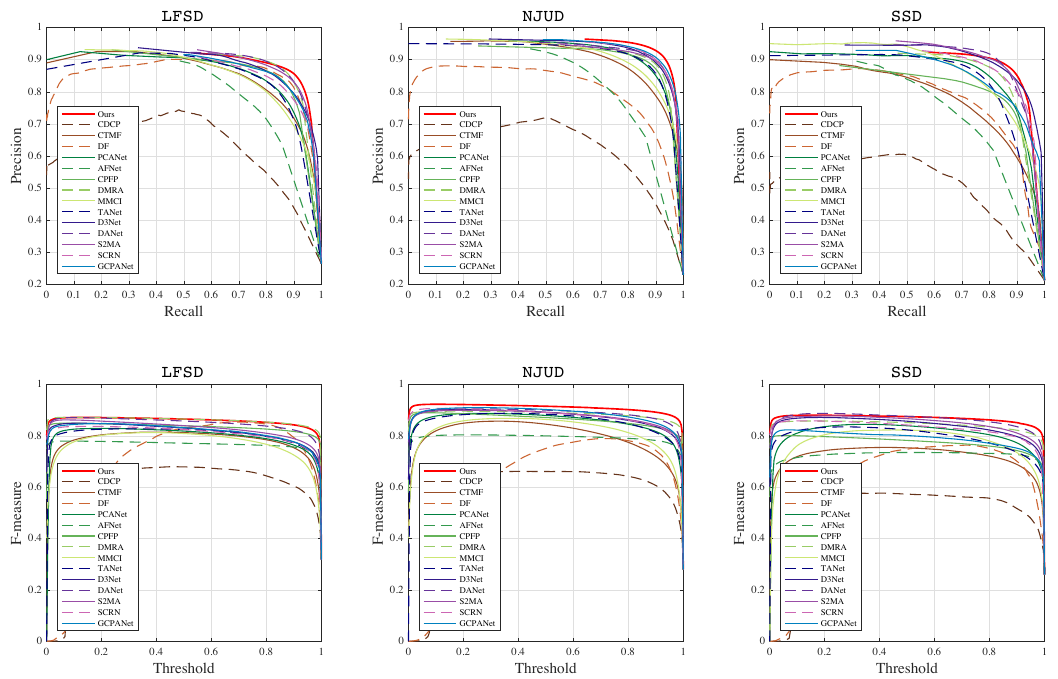} \\
			
		\end{minipage}
	}
\caption{PR curves and threshold curves of datasets SIP, NLPR, STEREO, LFSD, NJUD, SSD}.
		Precision (vertical axis) recall (horizontal axis) curves are shown in the 1st and 3rd rows.
		F-measure (vertical axis) threshold (horizontal axis) curves are shown in the 2nd and 4th rows.
    \label{PR}
\end{figure*}

Tab.~\ref{T-SOTA1} shows the results of the proposed model and 18 state-of-the-art SOD methods on 8 RGB-D datasets and also demonstrates that the ACFNet performs favorably against other state-of-the-art algorithms. Moreover, the PR curves (Fig.\ref{PR}) show that our approach outperform other methods.
The modle size is 83.76M. The computationl complexity is 33.07GMac.
Our network is a real-time fast network.
The speed of GRNet is 0.030s (33FPS), which is faster than most previous algorithms (CTMF (0.63s), MMCI (0.05s), TANet (0.07s), DMRA (0.06s), CPFP (0.17s), D$^{3}$Net (0.05s), DCF (0.032s), JL-DCP (0.111s), DANet (0.031s)).

\subsection{Qualitative Evaluation}
Fig.~\ref{compare} shows the visual examples produced by ACFNet and others.
From the top to the bottom, the utilization value of depth data is gradually decreasing.
Our algorithm accurately senses the change and dynamically adjusts the weights to make multiple types of features cooperate adaptively.
The proposed ACFNet performs better in various challenging scenarios, including complex background (1st, 2nd rows), fuzzy boundary (1st, 4th rows), inaccurate depth information (6th, 9th, 10th rows).

\subsection{Failure Cases}
\begin{figure}[htb]%
	\centering
	\includegraphics[width=1\columnwidth ]{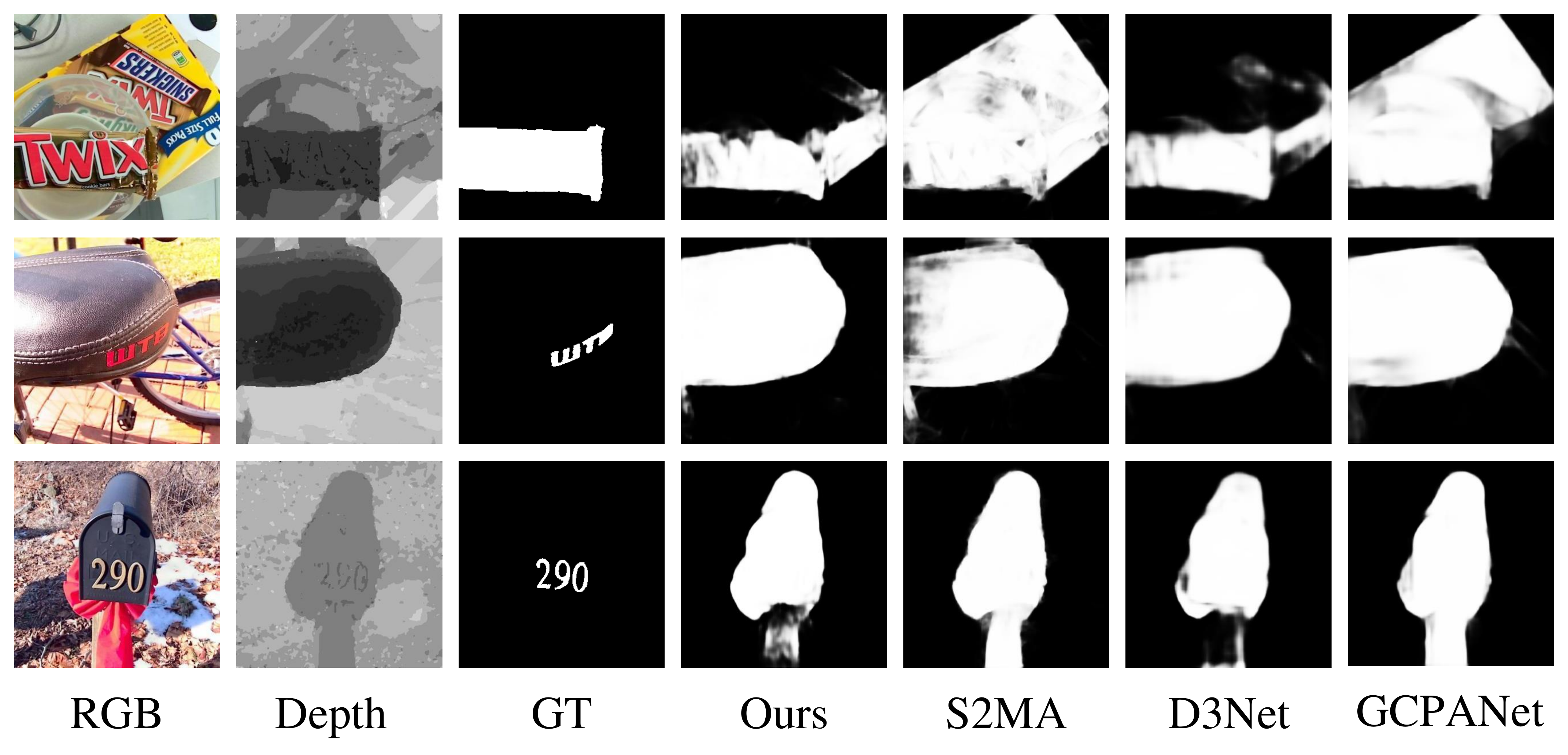}
	\caption{Failure examples.
	}
	\label{FCase}
\end{figure}

For future scholars to design better algorithms, we provide some failure cases in Fig.\ref{FCase}.
Compared with Ground Truth and saliency maps, it can be found that neither the RGB SOD method (GCPANet~\cite{2020-AAAI-GCPANet}) nor RGB-D SOD methods (S2MA~\cite{2020-CVPR-S2MA}, D3Net~\cite{2020-TNNLS-D3Net}) can locate the salient object accurately.
(1) Cluttered background area. When the background around the object is cluttered and complex, and the depth information is not accurate, our algorithm can not accurately segment the edge of the salient object, as shown in the 1st row.
(2) The depth map is invalid. When the depth map in the 2nd row is not helpful to find the object, our method can not accurately locate the text on the plane.
(3) Salient object is small. When the size of the 2nd and 3rd row objects are small, our algorithm ignores the small size salient objects.
Because the size of the shallowest feature adopted by our algorithm is 88, which improves the speed of the model, but it may be unfavorable for finding small objects.

\section{CONCLUSIONS}
In this paper, we propose a ResinRes network structure, which combines the advantages of early fusion and late fusion.
An adaptively cooperative semantic guidance (ACG) scheme is designed to facilitate the adaptive cooperation of different semantic features in the ResinRes network.
We further propose a type-based attention module (TAM) to optimize features. 
ACG and TAM optimize the transfer of feature streams according to the data attribute (RGB and Depth) and convolution attribute (normal convolutions and dilated convolutions) of features respectively.
Sufficient experiments conducted on 8 RGB-D SOD datasets demonstrate the superiority of our method.
In the future, we may add recurrent structure to ACFNet to optimize performance and apply it to the RGB-T task to verify its universality.
Besides, the ACG scheme may further improve its weight analysis accuracy through more complex structural design.

\bibliographystyle{plain}
\bibliography{temple}


%
%



\end{document}